\documentclass[letterpaper, 10 pt, conference]{ieeeconf}  

\IEEEoverridecommandlockouts                              

\usepackage{graphics} 
\usepackage{epsfig} 
\usepackage{mathptmx} 
\usepackage{amsmath} 
\usepackage{amssymb}  
\usepackage{lipsum}
\usepackage{bm}
\usepackage{xcolor}

\usepackage[style=ieee]{biblatex}

\DeclareSourcemap{
  \maps{
    \map{
      \pertype{article}
      \step[fieldset=language, null]
      \step[fieldset=url, null]
      \step[fieldset=doi, null]
      \step[fieldset=issn, null]
      \step[fieldset=isbn, null]
      \step[fieldset=note, null]
      \step[fieldset=editor, null]
      \step[fieldset=urldate, null]
      \step[fieldset=file, null]
    }
  }
}
\DeclareSourcemap{
  \maps{
    \map{
      \pertype{inproceedings}
      \step[fieldset=language, null]
      \step[fieldset=url, null]
      \step[fieldset=doi, null]
      \step[fieldset=issn, null]
      \step[fieldset=isbn, null]
      \step[fieldset=note, null]
      \step[fieldset=editor, null]
      \step[fieldset=urldate, null]
      \step[fieldset=file, null]
    }
  }
}
\DeclareSourcemap{
  \maps{
    \map{
      \pertype{incollection}
      \step[fieldset=language, null]
      \step[fieldset=url, null]
      \step[fieldset=doi, null]
      \step[fieldset=issn, null]
      \step[fieldset=isbn, null]
      \step[fieldset=note, null]
      \step[fieldset=editor, null]
      \step[fieldset=urldate, null]
      \step[fieldset=file, null]
    }
  }
}

\title{\LARGE \bf
Enhanced Capture Point Control Using Thruster Dynamics and QP-Based Optimization for Harpy
}

\author{Shreyansh Pitroda$^{1}$, Eric Sihite$^{2}$, Taoran Liu$^{1}$, Kaushik Venkatesh Krishnamurthy$^{1}$, \\ Chenghao Wang$^{1}$, Adarsh Salagame$^{1}$, Reza Nemovi$^{2}$, Alireza Ramezani$^{1*}$, and Morteza Gharib$^{2}$
\thanks{$^{1}$ The authors are with SiliconeSynapse Labs, the Department of Electrical Engineering, Northeastern University, USA.}%
\thanks{$^{2}$ The authors are with the Department of Aerospace Engineering, California Institute of Technology, USA.}%
\thanks{$^{*}$ The corresponding author. Email: a.ramezani@northeastern.edu}%
}

\begin{document}

\maketitle
\thispagestyle{empty}
\pagestyle{empty}

\begin{abstract}

Our work aims to make significant strides in understanding unexplored locomotion control paradigms based on the integration of posture manipulation and thrust vectoring. These techniques are commonly seen in nature, such as Chukar birds using their wings to run on a nearly vertical wall. In this work, we developed a capture-point-based controller integrated with a quadratic programming (QP) solver which is used to create a thruster-assisted dynamic bipedal walking controller for our state-of-the-art Harpy platform. Harpy is a bipedal robot capable of legged-aerial locomotion using its legs and thrusters attached to its main frame. While capture point control based on centroidal models for bipedal systems has been extensively studied, the use of these thrusters in determining the capture point for a bipedal robot has not been extensively explored. The addition of these external thrust forces can lead to interesting interpretations of locomotion, such as virtual buoyancy studied in aquatic-legged locomotion. In this work, we derive a thruster-assisted bipedal walking with the capture point controller and implement it in simulation to study its performance.

\end{abstract}


\section{Introduction}

Raibert's robots \cite{murphy_littledog_2011,} and those from Boston Dynamics \cite{noauthor_robots_nodate,boston_dynamics_all_2024,boston_dynamics_meet_2024} stand out as some of the most successful legged robots, capable of robust hopping or trotting even amid substantial unplanned disturbances. In addition to these advancements, various underactuated and fully actuated bipedal robots have emerged \cite{park_finite-state_2013,buss_preliminary_2014,ramezani_performance_2014,grizzle_progress_nodate}. Agility Robotics' Cassie \cite{apgar_fast_2018}, ASIMO \cite{shigemi_asimo_2019}, HRP \cite{kajita_biped_2003}, HUBO \cite{wang_drc-hubo_2014} and TALOS \cite{stasse_talos_2017} display abilities such as walking, running, dancing, and navigating stairs, while Atlas can recover from pushes \cite{koolen_design_2016}.

Boston Dynamics' Atlas is known for performing acrobatics such as jumping and flipping, parkour, running and bounding, and gymnastic routines. However, despite these capabilities, all these systems remain susceptible to falling and have been achieved through extensive trial and error, much of which is not publicized. Consequently, it is difficult to assess the repeatability or robustness of these maneuvers in real-world applications. 

\begin{figure}[t]
    \vspace{0.05in}
    \centering
    \includegraphics[width=0.8\linewidth]{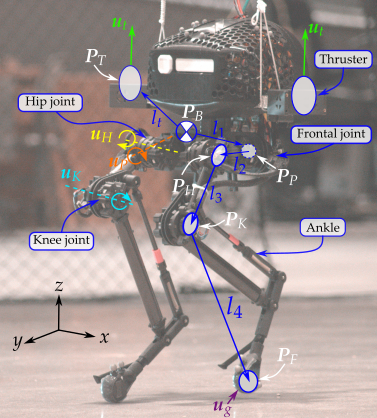}
    \caption{Illustrates the Harpy platform, which motivates our thruster-assisted dynamic legged locomotion.}
    \label{fig:cover-image}
    \vspace{-0.05in}
\end{figure}

Even humans, renowned for their natural, dynamic, and robust gaits, cannot consistently recover from severe terrain perturbations, external pushes, or slips on icy surfaces. Our objective is to enhance the robustness of these systems by implementing a distributed array of thrusters and employing nonlinear control techniques.

The primary contribution of this work is the development of a capture-point-based controller integrated with a quadratic programming (QP) solver. This combination is used to create a thruster-assisted dynamic bipedal walking controller for our state-of-the-art Harpy platform, as illustrated in Fig.~\ref{fig:cover-image}. 

Recent tests have explored the application of thrusters (i.e., thrust vectoring) and posture manipulation in notable robots such as the Multi-modal mobility morphobot (M4) \cite{sihite_multi-modal_2023, sihite_efficient_2022, mandralis_minimum_2023} and LEONARDO \cite{kim_bipedal_2021, liang_rough-terrain_2021, sihite_optimization-free_2021, sihite_efficient_2022}. 

The M4 robot aims to enhance locomotion versatility by combining posture manipulation and thrust vectoring, enabling modes such as walking, wheeling, flying, and loco-manipulation. LEONARDO, a quadcopter with two legs, can perform both quasi-static walking and flying. However, neither of these robots fully demonstrate dynamic legged locomotion and aerial mobility. Integrating these modes poses a significant challenge due to conflicting requirements (see \cite{sihite_multi-modal_2023}). Achieving both dynamic walking and aerial mobility within a single platform remains a major obstacle in hardware and control design.

Our work aims to make significant strides in understanding unexplored locomotion control paradigms based on the integration of posture manipulation and thrust vectoring. These techniques, commonly used by birds, which are known for their locomotion plasticity and robust locomotion feats. For instance, Chukar birds can perform a wing-assisted incline running (WAIR) maneuver \cite{dial_wing-assisted_2003-1, tobalske_aerodynamics_2007}. In the WAIR maneuver, Chukar birds utilize their flapping wings and the resulting aerodynamic forces to increase contact forces, allowing them to ascend steep slopes that conventional bipedal robots would find challenging to navigate.

In this study, we employ a detailed model of Harpy (depicted in Fig.~\ref{fig:cover-image}) using Matlab Simscape to evaluate our controller's effectiveness. Harpy is equipped with eight custom-designed high-energy density actuators for dynamic walking, along with electric ducted fans mounted on its torso sides. Harpy's height measures 600 cm and weighs 4.5 Kg. It hosts a computer based on Elmo amplifiers for real-time low-level control command executions.

Harpy's design integrates advantages from both aerial and dynamic bipedal legged systems. Currently, the hardware design and assembly of Harpy have been completed~\cite{pitroda_dynamic_2023}, and our primary goal is to explore various control design strategies for this platform \cite{dangol_feedback_2020-1, dangol_control_2021}.

While capture point control based on centroidal models for bipedal systems has been extensively studied \cite{koolen_design_2016,ramos_generalizations_2015,hong_capture_2019,bickel_capture_2009}, the incorporation of thruster forces that can influence the dynamics of linear inverted pendulum models, often used in capture point-based works, has not been explored before and is limited to \cite{iqbal_extended_2021,krause_stabilization_2012,pratt_capture_2006}. The inclusion of these external thrust forces can lead to interesting interpretations of locomotion, such as virtual buoyancy studied in aquatic-legged locomotion.

In this work, we aim to design a planning and control approach that enables stable trotting in place by extending our previous work~\cite{pitroda_capture_2024}. The main contributions of this paper are: a) formulating a QP-based reference tracking controller which has dynamics of VLIP model and capture point. b) To study the effect of thruster force on the stability and robustness of the system, we made thruster force a parameter for the QP controller. c) we want to take some meaningful steps towards exploring the unexplored domain of thruster-assisted dynamic terrestrial locomotion.

This work is structured as follows: we present the derivations of Harpy reduced order model, followed by the capture point control,Online planning using Quadratic Programming(QP), simulation results, and concluding remarks.

\section{Thruster-Assisted Bipedal Walking Model}

\begin{figure}[t]
    \centering
    \vspace{0.05in}
    \includegraphics[width=1.0\linewidth]{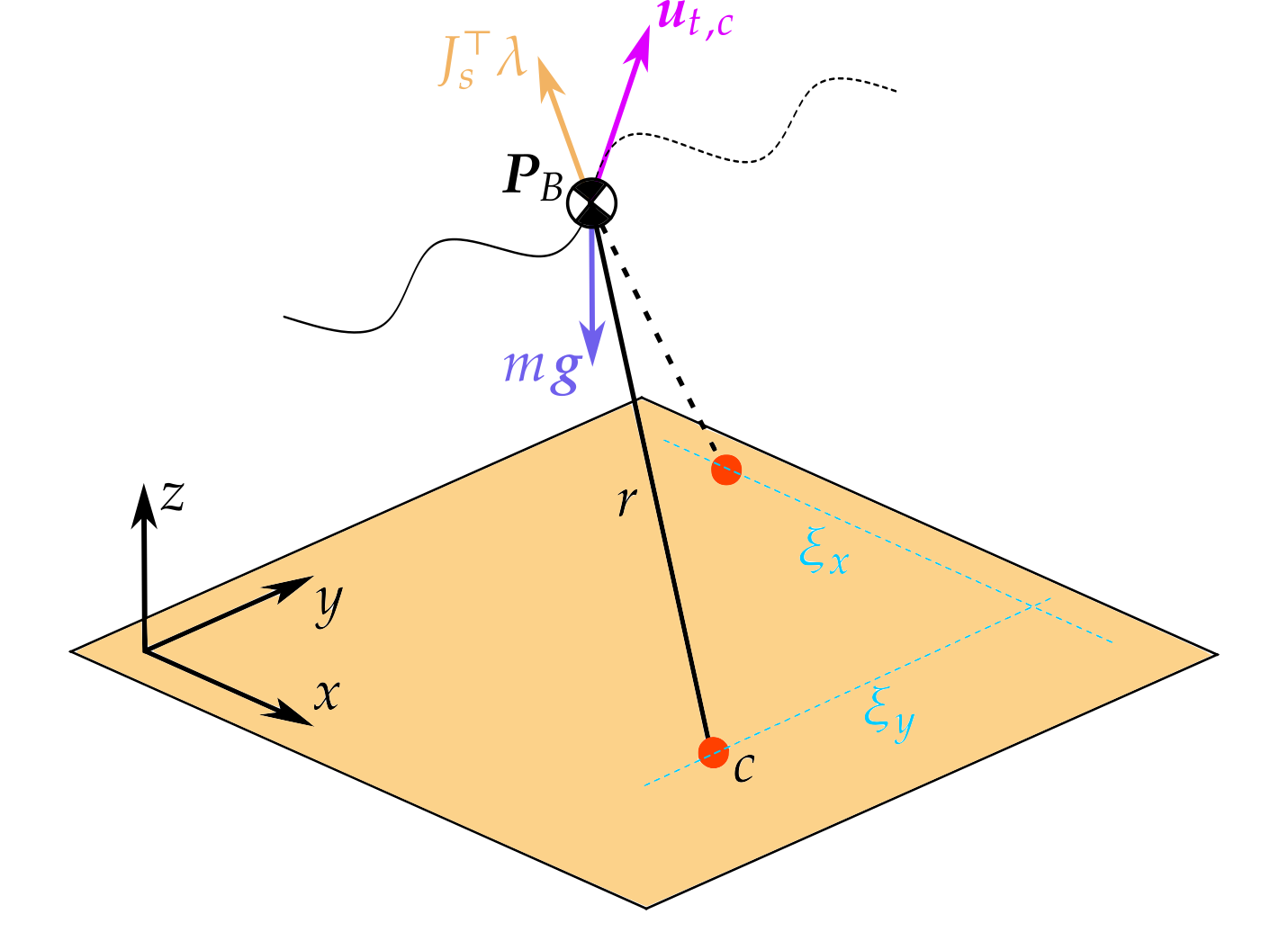}
    \caption{Shows the Harpy reduced-order model with thrusters, represented by a single body and massless legs.}
    \label{fig:hrom}
    \vspace{-0.05in}
\end{figure}

This section outlines the dynamics formulation of the robot which is used in the numerical simulation in Section 3, in addition to the reduced order models which are used in the controller design. Figure \ref{fig:cover-image} shows the kinematic configuration of Harpy which listed the center of mass (CoM) positions of the dynamic components, joint actuation torques, and thruster torques. The system model has a combined total of 12 degrees-of-freedoms (DoFs): 6 for the body and 3 on each leg. Due to the symmetry, the left and right side of the robot follow a similar derivations so only the general derivations are provided in this section.

\subsection{Energy-based Lagrange Formalism}

The Harpy equations of motion are derived using Euler-Lagrangian dynamics formulation. In order to simplify the system, each linkage is assumed to be massless, with the mass concentrated at the body and the joint motors. Consequently, the lower leg kinematic chain is considered massless, significantly simplifying the system. The three leg joints are labeled as the hip frontal (pelvis $P$), hip sagittal (hip $H$), and knee sagittal (knee $K$), as illustrated in Fig. \ref{fig:cover-image}. The thrusters are also considered massless and capable of providing forces in any direction to simplify the problem.

Let $\gamma_h$ be the frontal hip angle, while $\phi_h$ and $\phi_k$ represent the sagittal hip and knee angles, respectively. The superscripts $\{B,P,H,K\}$ represent the frame of reference about the body, pelvis, hip, and knee, while the inertial frame is represented without the superscript. Let $R_B$ be the rotation matrix from the body frame to the inertial frame (i.e., $\bm x = R_B\, \bm x^B$). The pelvis motor mass is added to the body mass. Then, the positions of the hip and knee centers of mass (CoM) are defined using kinematic equations:
\begin{equation}
\begin{gathered}
    \bm{p}_P = \bm{p}_{B} + R_{B}\, \bm{l}_{1}^{B}, \\
    \bm{p}_H = \bm{p}_{P} + R_{B}\,R_x(\gamma_h)\, \bm{l}_{2}^{P} \\
    \bm{p}_K = \bm{p}_{H} + R_{B}\,R_x(\gamma_h)\,R_y(\phi_h) \bm{l}_{3}^{H},
\end{gathered}
\label{eq:pos_com}
\end{equation}
where $R_x$ and $R_y$ are the rotation matrices about the $x$ and $y$ axes, respectively, and $\bm l$ is the length vector representing the configuration of Harpy, which remains constant in its respective local frame of reference. The positions of the foot and thrusters are defined as:
\begin{equation}
\begin{gathered}
    \bm{p}_F = \bm{p}_{K} + R_{B}\,R_x(\gamma_h)\,R_y(\phi_h)\,R_y(\phi_k)\, \bm{l}_{4}^{K} \\
    \bm{p}_T = \bm{p}_{B} + R_{B}\, \bm{l}_{t}^{B}
\end{gathered}
\label{eq:pos_other}
\end{equation}
where the length vector from the knee to the foot is $\bm l_4^K = [-l_{4a}\cos{\phi_k}, 0, -( l_{4b} + l_{4a}\sin{\phi_k})]^\top$, which represents the kinematic solution to the parallel linkage mechanism of the lower leg. Let $\bm \omega_B$ be the angular velocity of the body. Then, the angular velocities of the hip and knee are defined as: $\bm \omega_H^B = [\dot{\gamma}_h,0,0]^\top + \bm \omega_B^B$ and $\bm \omega_K^H = [0,\dot{\phi}_h,0]^\top + \bm \omega_H^H$. Consequently, the total energy of Harpy for the Lagrangian dynamics formulation is defined as follows:
\begin{equation}
\begin{aligned}
    K &= \tfrac{1}{2} \textstyle \sum_{i \in \mathcal{F}} \left( 
        m_i\,\bm p_i^\top\, \bm p_i + 
        \bm \omega_i^{i \top} \, \hat I_i \, \bm \omega_i^i \right) \\
    V &= - \textstyle \sum_{i \in \mathcal{F}} \left( 
        m_i\,\bm p_i^\top\, [0,0,-g]^\top \right),
\end{aligned}
\label{eq:energy}
\end{equation}
where $\mathcal{F} = \{B,H_L,K_L,H_R,K_R\}$ represents the relevant frames of reference and mass components (body, left hip, left knee, right hip, right knee), and the subscripts $L$ and $R$ denote the left and right sides of the robot, respectively. Furthermore, $\hat I_i$ denotes the inertia about its local frame, and $g$ is the gravitational constant. This constitutes the Lagrangian of the system, given by $L = K - V$, which is utilized to derive the Euler-Lagrange equations of motion. The dynamics of the body's angular velocity are derived using the modified Lagrangian for rotation in $SO(3)$ to avoid using Euler angles and the potential gimbal lock associated with them. This yields the following equations of motion following Hamilton's principle of least action:
\begin{equation}
\begin{gathered}
    \tfrac{d}{dt}\left( \tfrac{\partial L}{\partial \bm \omega_B^B}  \right) + 
    \bm \omega_B^B \times \tfrac{\partial L}{\partial \bm \omega_B^B} + 
    \textstyle \sum_{j=1}^{3} \bm r_{Bj} \times \tfrac{\partial L}{\partial \bm r_{Bj}} = \bm u_1, \\
    \tfrac{d}{dt}\left( \tfrac{\partial L}{\partial \dot {\bm q}}  \right) - 
    \tfrac{\partial L}{\partial \bm q} = \bm u_2, \\ 
    \tfrac{d}{dt} R_B = R_B\, [\bm \omega_B^B]_\times,
\end{gathered}
\label{eq:eom_eulerlagrange}
\end{equation}
where $[\, \cdot \, ]_\times$ denotes the skew symmetric matrix, $R_B^\top = [\bm r_{B1}, \bm r_{B2}, \bm r_{B3}]$, $\bm q = [\bm p_B^\top, \gamma_{h_L}, \gamma_{h_R}, \phi_{h_L}, \phi_{h_R}]^\top$ represents the dynamical system states other than $(R_B,\bm \omega^B_B)$, and $\bm u$ denotes the generalized forces. The knee sagittal angle $\phi_k$, which is not associated with any mass, is updated using the knee joint acceleration input $\bm u_k = [\ddot{\phi}_{k_L}, \ddot{\phi}_{k_R}]^\top$. Then, the system acceleration can be derived as follows:
\begin{equation}
\begin{gathered}
    M \bm a + \bm h = B_j\, \bm u_j + B_t\, \bm u_t + B_g\, \bm u_g
\end{gathered}
\label{eq:eom_accel}
\end{equation}
where $\bm a = [ \dot{\bm \omega}_B^{B\top}, \ddot{\bm q}^\top, \ddot{\phi}_{k_L}, \ddot{\phi}_{k_R}]^\top$, $\bm u_t$ denotes the thruster force, $\bm u_j = [u_{P_L}, u_{P_R}, u_{H_L}, u_{H_R}, \bm u_k^\top]^\top$ represents the joint actuation, and $\bm u_g$ stands for the ground reaction forces (GRFs). The variables $M$, $\bm h$, $B_t$, and $B_g$ are functions of the full system states:
\begin{equation}
    \bm x = [\bm r_{B}^\top, \bm q^\top, \phi_{K_L}, \phi_{K_R}, \bm \omega_B^{B \top}, \dot{\bm q}^\top, \dot{\phi}_{K_L}, \dot{\phi}_{K_R}]^\top,
\label{eq:states}
\end{equation}
where the vector $\bm r_B$ contains the elements of $R_B$. Introducing $B_j = [0_{6 \times 6}, I_{6 \times 6}]$ allows $\bm u_j$ to actuate the joint angles directly. Let $\bm v = [\bm \omega_B^{B\top}, \dot{\bm q}^\top]^\top$ denote the velocity of the generalized coordinates. Then, $B_t$ and $B_g$ can be defined using the virtual displacement from the velocity as follows:
\begin{equation}
\begin{aligned}
    B_t = \begin{bmatrix}
        \begin{pmatrix}
        \partial \dot{\bm p}_{T_L} / \partial \bm v \\
        \partial \dot{\bm p}_{T_R} / \partial \bm v
        \end{pmatrix}^\top
        \\
        0_{2 \times 6}
    \end{bmatrix}, \quad
    B_g = \begin{bmatrix}
        \begin{pmatrix}
        \partial \dot{\bm p}_{F_L} / \partial \bm v \\
        \partial \dot{\bm p}_{F_R} / \partial \bm v
        \end{pmatrix}^\top
        \\
        0_{2 \times 6}
    \end{bmatrix}.
\end{aligned}
\label{eq:generalized_forces}
\end{equation}
The vector $\bm u_t = [\bm u_{t_L}^\top, \bm u_{t_R}^\top]^\top$ is composed of the left and right thruster forces $\bm u_{t_L}$ and $\bm u_{t_R}$, respectively. 



The full-dynamics model can be derived using equations \eqref{eq:eom_eulerlagrange} to \eqref{eq:generalized_forces} to form $\dot{\bm x} = \bm f(\bm x, \bm u_j, \bm u_t, \bm u_g)$. Finally, using the full-dynamics derived above, we proceed to ROM derivations. As shown in Fig. \ref{fig:hrom}, the model is described using the inverted pendulum model, where the length of $r$ can be adjusted through the change in leg conformation, i.e., variable-length inverted pendulum model (VLIP).

\begin{figure}[t]
    \centering
    \vspace{0.05in}
    \includegraphics[width=1\linewidth]{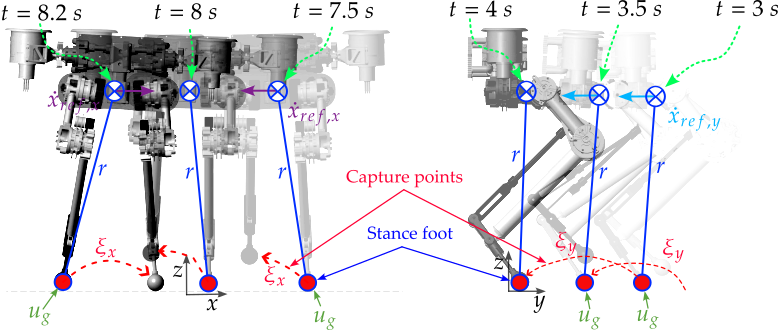}
    \caption{Snapshots of Harpy dynamically trotting on flat ground.}
    \label{fig:Harpy_walking}
    \vspace{-0.05in}
\end{figure}

\section{Thruster-assisted Locomotion Controls}

\subsection{Capture Point Derivations}

To design the controller, we simplify the dynamics of VLIP as depicted in Fig.~\ref{fig:hrom} by projecting it onto sagittal and frontal plane. We start with deriving the equation of motion for sagittal plane.
\begin{equation}
\begin{aligned}
& m \ddot{p}_{B,x}=|\bm\lambda| \sin \theta_L+\left|\bm u_{t, c}\right| \sin \theta_T \\
& m \ddot{p}_{B,z}=-mg+|\bm \lambda| \cos \theta_L+\left|\bm u_{t, c}\right| \cos \theta_T
\end{aligned}
\label{eq:sagittal-vlip}
\end{equation}
where $\theta_L$ is the angle made by the pendulum with vertical in saggital plane and $\theta_T$ is the thruster angle with respect to vertical.The linear pendulum model can be enforced by setting $p_{B,z}=z_0$ and $\ddot p_{B,z}=0$. Therefore, the magnitude of $\bm \lambda$ is determined by

\begin{equation}
    \left|\bm \lambda\right|=\left(mg-\left|\bm u_{t, c}\right| \cos \theta_T\right) \frac{\left|\bm r\right|}{z_0}
\end{equation}
By substituting $\sin \theta_L=\frac{x}{r}$ and $\left|\bm \lambda\right|$ from above into Eq.~\ref{eq:sagittal-vlip}, $\ddot p_{B,x}$ is given by
\begin{equation}
    m\ddot{p}_{B,x}=\frac{x}{z_0}\left(mg-\left|\bm u_{t, c}\right| \cos \theta_T\right)+\left|\bm u_{t, c}\right| \sin \theta_T
\end{equation}
Note that if through torso angle manipulation thruster actions around the CoM $\bm u_{t,c}$ are kept perpendicular to the ground surface, i.e., $\theta_T=0$, then we can express the virtual mass-spring model with a negative stiffness rate $-\left(g-\frac{\left|\bm u_{t,c}\right|}{m}\right)$ as follows:
\begin{equation}
    \ddot{p}_{B,x}=\left(g-\frac{\left|\bm u_{t,c}\right|}{m}\right) \frac{p_{B,x}}{z_0}
\label{eq: sagittal dynamics-vlip}
\end{equation}
Similarly, we can derive the equation of motion for frontal plane.
\begin{equation}
    \ddot{p}_{B,y}=\left(g-\frac{\left|\bm u_{t,c}\right|}{m}\right) \frac{p_{B,y}}{z_0}
\label{eq: frontal dynamics-vlip}
\end{equation}
Since the stiffness rate in this model is negative and dictated by the thrusters, we refer to this model as virtual buoyancy. It is possible to observe that the thruster force can reduce the walking frequency, similar to submersed aquatic-legged locomotion. The orbital energy $E$ of the virtual buoyancy model is given by 
\begin{equation}
    E=\frac{1}{2} \dot{p}_{B,x}^2-\frac{1}{2}\left(g-\frac{\left|\bm u_{t, c}\right|}{m}\right) \frac{p_{B,x}^2}{z_0}
\end{equation}
When the CoM moves towards the foot and $E > 0$, there is sufficient energy for the CoM to pass over the foot and maintain its motion. Conversely, if $E < 0$, the CoM halts and changes direction before reaching over the foot. At $E = 0$, the CoM comes to a rest directly above the foot. This equilibrium state, $E = 0$, defines the two eigenvectors of the buoyancy model, expressed as:
\begin{equation}
    \dot{p}_{B,x}= \pm p_{B,x} \sqrt{\frac{g-\frac{\left|\bm u_{t,c}\right|}{m}}{z_0}}
\end{equation}
The equation above depicts a saddle point characterized by one stable and one unstable eigenvector. In the stable eigenvector, $p_{B,x}$ and $\dot p_{B,x}$ exhibit opposite signs, indicating that the CoM is approaching the CoP. Conversely, in the unstable eigenvector, they share the same signs, indicating that the CoM is moving away from the CoP. The orbital energy of the inverted pendulum remains constant until the swing leg is placed and the roles of the feet are exchanged. Assuming this exchange occurs instantaneously without energy loss, we can determine the foot placement based on the capture point, given by
\begin{equation}
\begin{aligned}
   & p_{B,x}=\dot p_{B,x} \sqrt{\frac{z_0}{g-\frac{\left|\bm u_{t,c}\right|}{m}}}
   \end{aligned}
\label{eq:Capture-point-sagittal}
\end{equation}
Similarly, we can find the capture point for frontal dynamics.
\begin{equation}
\begin{aligned}
   & p_{B,y}=\dot p_{B,y} \sqrt{\frac{z_0}{g-\frac{\left|\bm u_{t,c}\right|}{m}}}
\end{aligned}
\label{eq:Capture-point-frontal}
\end{equation}

\subsection{Capture Point Online Planning Using Quadratic Programming}

The objective of the controller is to generate the desired foot placement location such that the robot follows a certain velocity. Capture point $\left( p_{B,x}, p_{B,y} \right)$ is further extended to allow the robot to follow a certain velocity. For simplicity, the desired motion is specified by a constant base velocity $\dot{\bm{x}}_{ref} = \left( \dot x_{ref,x}, \dot x_{ref,y} \right)$. The capture point corresponding to desired velocity is given by 
\begin{equation}
\begin{aligned}
    \xi_{x} &=K_{x}\left( \dot p_{B,x} - \dot x_{ref,x} \right) \sqrt{\frac{z_0}{g-\frac{\left|\bm u_{t,c}\right|}{m}}} \\
    \xi_{y} &=K_{y}\left( \dot p_{B,y} - \dot x_{ref,y} \right) \sqrt{\frac{z_0}{g-\frac{\left|\bm u_{t,c}\right|}{m}}}
\end{aligned}
\label{eq: desired-Capture-point}
\end{equation}
Here $K_{x},K_{y}$ is gain and can be tuned to see how much the COM velocity affects the desired foot location. This will allow us to consider the hardware limitation. Similar concept was used in \cite{iqbal_extended_2021}.we can differentiate the equation \eqref{eq: desired-Capture-point} to find the change in capture point and substitute equations \eqref{eq: sagittal dynamics-vlip} and \eqref{eq: frontal dynamics-vlip} to get,
\begin{equation}
\begin{aligned}
   & \dot \xi_{x}=K_{x}\left(\omega \right)p_{B,x}\\
   & \dot \xi_{y}=K_{y}\left(\omega \right)p_{B,y}
\label{eq:dot_xi}
\end{aligned}
\end{equation}
where $\omega = \sqrt{\left(g -\frac{\left|\bm u_{t,c}\right|}{m}\right)\frac{1}{z_0}}$ is the natural frequency of the Vlip model. Above equation is unstable and response is governed by $\omega$. Using $\bm u_{t,c}$ we can make system sinusodal and can change frequency of walking.

\begin{figure}[t]
    \centering
    \vspace{0.05in}
    \includegraphics[width=1\linewidth]{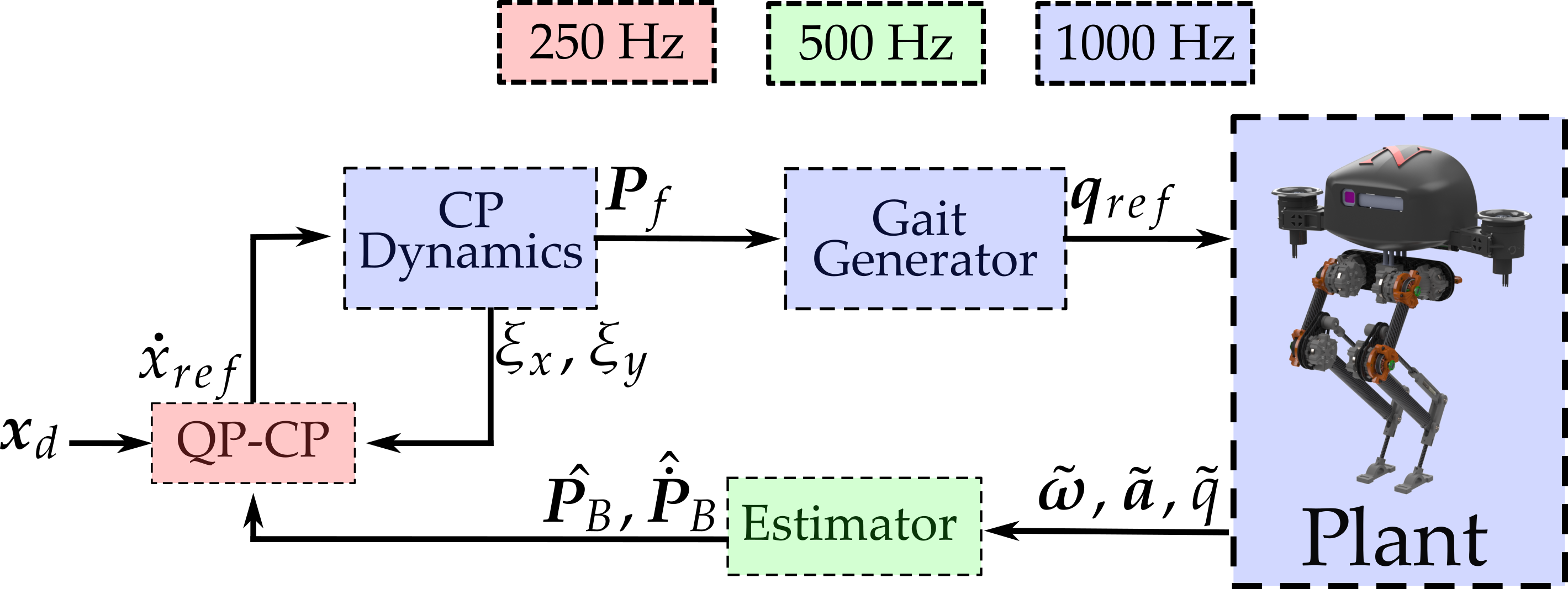}
    \caption{Harpy controller architecture.}
    \label{fig:Control-architecture}
    \vspace{-0.05in}
\end{figure}

From equations \eqref{eq:dot_xi} \eqref{eq:Capture-point-frontal} \eqref{eq:Capture-point-sagittal}, we can find the state space model for QP.
\begin{equation}
\begin{aligned}
\begin{bmatrix}
    \dot{\bm p}_{B} \\
    \dot{\bm \xi}
\end{bmatrix} = 
\begin{bmatrix}
 0 & \omega\\
 K \omega & 0
\end{bmatrix} 
\begin{bmatrix}
    \bm p_{B} \\
    \bm \xi
\end{bmatrix} + 
\begin{bmatrix}
    1 \\
    0
\end{bmatrix}
\dot{\bm x}_{ref}
\label{eq:state-space-model}
\end{aligned}
\end{equation}
Here, $\bm p_{B} = \left[p_{B,x},p_{B,y}\right]^\top$ and $\bm \xi = \left[ \xi_x,\xi_y\right]^\top$. We can further discretize the dynamics for Qp formulation.
\begin{equation}
   \bm x_{K+1} = A_{d}\bm x_{K} + B_{d}\bm u_{K}
\end{equation}
We can eliminate the states from the decision variables by expressing them as function of current state $x_{0}$ and control efforts
\begin{equation}
   \bm x = F\bm x_{0} + G\bm u
\end{equation}
where
\begin{equation}
\begin{aligned}
   \bm x &= \left[x^\top_{0} x^\top_{1} x^\top_{2} \hdots x^\top_{N}\right]^\top \\
   \bm u &= \left[u^\top_{0} u^\top_{1} u^\top_{2} \hdots u^\top_{N-1} \right]^\top
\end{aligned}
\end{equation}
\begin{gather}
   F = 
   \begin{bmatrix}
   \mathbb{I} \\
   A_d \\
   \vdots \\
   A^{N}_d
   \end{bmatrix}
   G = 
    \begin{bmatrix}
    0 &  &   &  & \\
    B_d & 0 &  &  & \\
    \vdots & & \ddots & & \\
    A^{N-1}_dB_d & A^{N-2}_dB_d & \hdots &A_d B_d & B_d\\
    \end{bmatrix}
\end{gather}
We create error tracking QP formulation with decision variable as $\bm u$.
\begin{equation}
   \min_{\bm u}~\bm u^\top P\bm u + c^\top \bm u
\label{eq:qp-formulation}
\end{equation}
subject to
\begin{equation}
\begin{aligned}
    A_{in}\bm u < B_{in} \\
\end{aligned}
\label{eq:qp-constraints}
\end{equation}
%
%
%
We have constrained QP with inequality constraints and dense matrices, where $P$ and $c$ are defined as follows:
\begin{align*}
    P & = G^\top QG + R\\
    c & = \left(x_{0}-x_{d,0}\right) F^\top QG \\
\label{eq:qp-matrices}
\end{align*}
In this case, $\bm x_{0} $ are the initial states and $\bm x_{d}$ are the reference for those states. $Q$ is cost associated with error $\left(\bm x-\bm x_{d}\right)$ and $R$ is cost for control effort $\bm u$

\section{Results and Discussion}

\begin{figure}[t]
\centering
\includegraphics[width=1\linewidth]{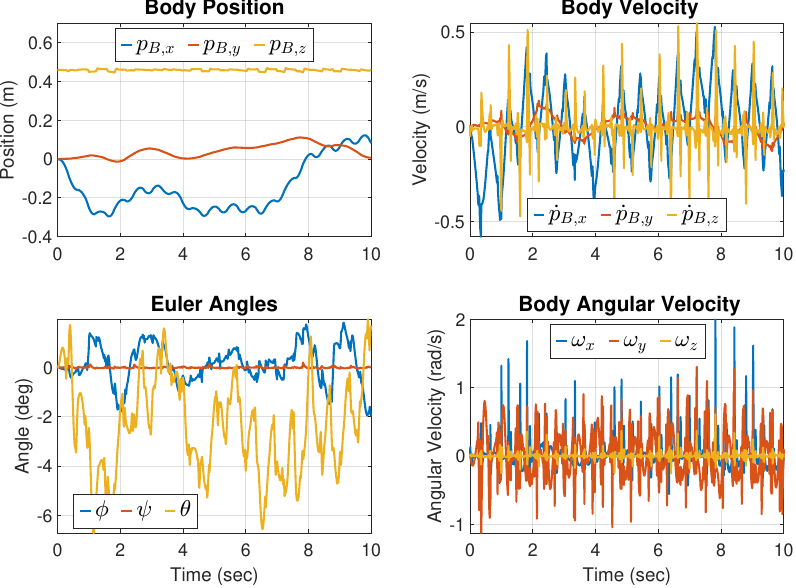}
    \caption{Illustrates Harpy states.}
    \label{fig:states}
\end{figure}

\begin{figure}[t]
    \centering
    \includegraphics[width=1\linewidth]{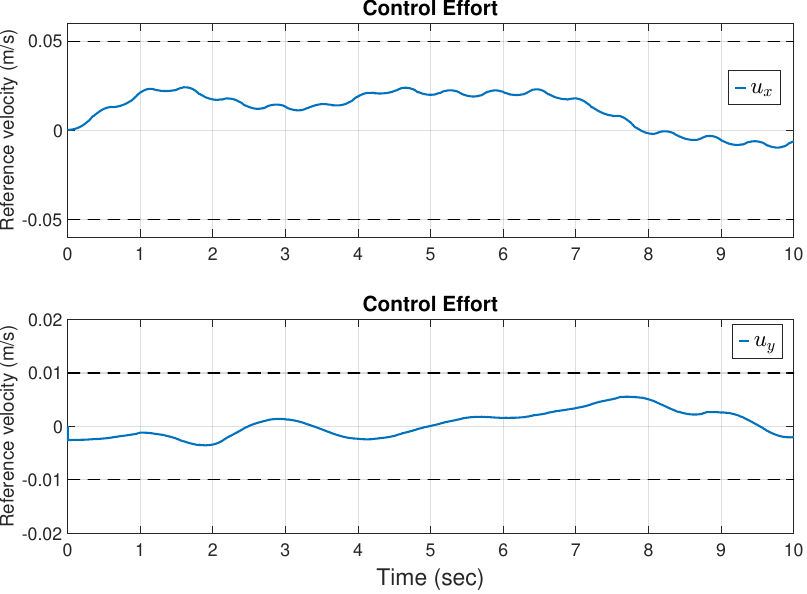}
    \caption{Illustrates Control effort generated by the QP controller. Dashed lines represent the bounds for $\bm u$.}
    \label{fig:Control-Effort}
\end{figure}

\begin{figure}[t]
    \centering
    \includegraphics[width=1\linewidth]{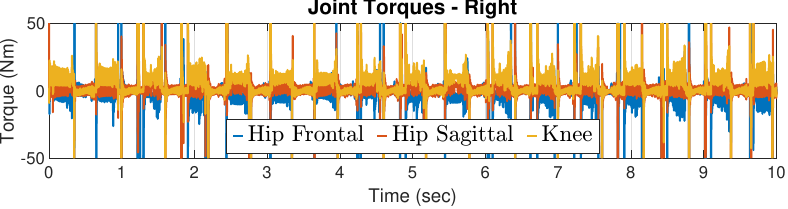}
    \caption{Illustrates Harpy's right foot joint torques.}
    \label{fig:Joint-Torque}
\end{figure}

\begin{figure}[t]
    \centering
    \includegraphics[width=1\linewidth]{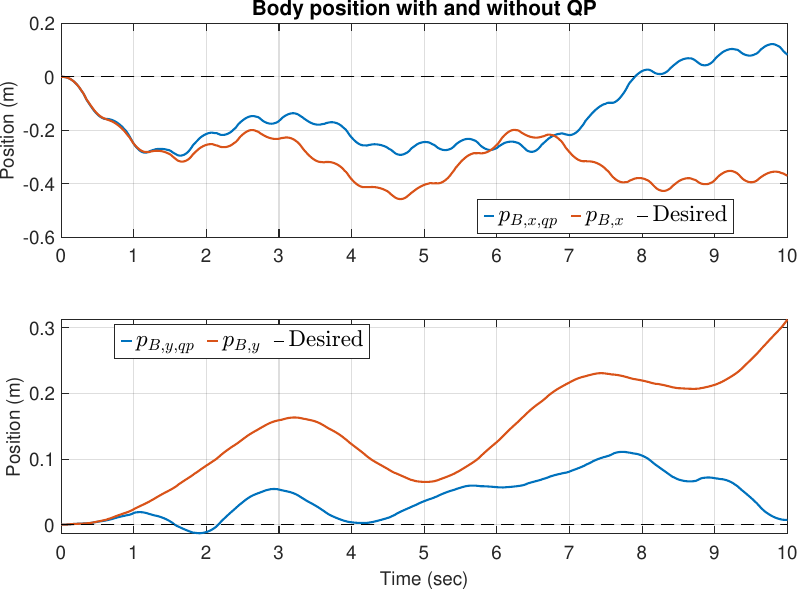}
    \caption{Illustrates the body position with and without QP controller. The position states are more unstable without QP.}
    \label{fig:BodyPosCompare}
\end{figure}

\begin{figure}[t]
    \centering
    \includegraphics[width=1\linewidth]{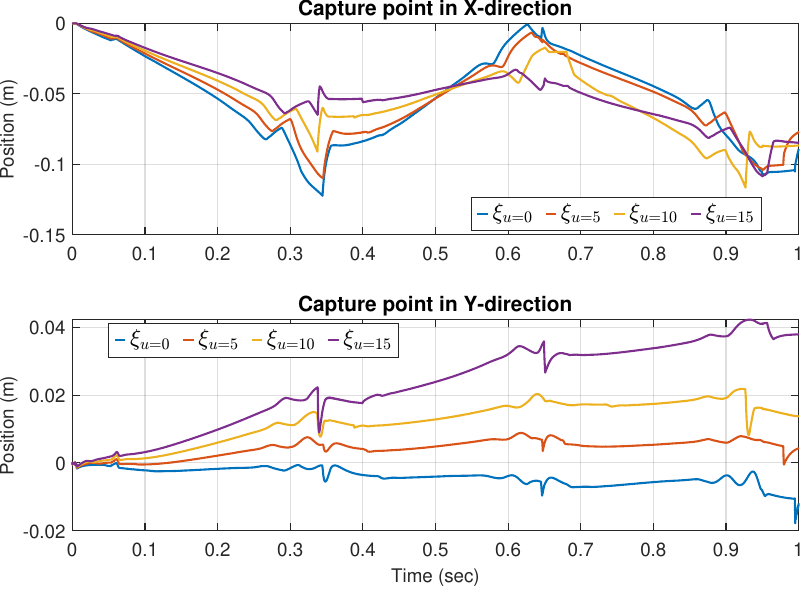}
    \caption{Illustrates the impact of varying thrust values on the capture point.}
    \label{fig:CP_compare}
\end{figure}

This section presents MATLAB simulation results (depicted in Fig. \ref{fig:Harpy_walking}) demonstrating  the efficacy of the controller algorithm. Figure~\ref{fig:Control-architecture} illustrates the overall controller implemented in Simulink. The condensed Quadratic Programming (QP) formulation, depicted in \eqref{eq:qp-formulation}, utilizes the qpSWIFT package \cite{pandala_qpswift_2019}. The weighting matrices in the cost function are defined as $ Q = diag\left(100,100,0.1,0.1\right)$ and $ R = diag\left(50,55\right)$. The gain values for the capture point dynamics \eqref{eq: desired-Capture-point} are set to $K_{x} = 1$ and $K_{y} = 1$ and reference for states is $\bm x_d = \left[0,0,0,0\right]^\top$. 

Figure \ref{fig:states} presents the states of the robot, demonstrating mild variations in body velocity and orientation, indicating successful stabilization by the proposed controller. Figures
\ref{fig:Control-Effort} and \ref{fig:Joint-Torque} show that the control efforts and joint torques are well within the desired bounds. Figure \ref{fig:BodyPosCompare} compares the body position with and without the QP controller. Without QP control, the robot exhibits stability but continuous drift. On the other hand, with QP control, the robot stabilizes towards the reference positions (dotted lines), highlighting the effectiveness of our QP-CP controller. 

Figure \ref{fig:CP_compare} demonstrates the effect of thrust force on the capture point. $u$ is the combined thruster force generated by both the thrusters. it could be seen from the top graph that as we increase the thruster force stability of the robot increases resulting in smaller capture point. Bottom graph shows as we increase the thruster force the capture point length increases. This is attributed to the position of the thruster being   fixed on the robot and as the robot falls, it gives additional acceleration in that direction.

These results show the robustness and efficacy of the proposed controller in stabilizing the robot and maintaining desired trajectories.

\section{Conclusions and Future Work}

We presented the design and implementation of the controller which uses the capture point dynamics and QP-based reference tracking for stable trotting on the Harpy platform. we make the thruster force generated by thrusters on Harpy as a parameter in the QP controller. Additionally, we show a detailed analysis of how the thrust force affects the stability of the robot. 

Future work will focus on improving the controller tracking performance by adding ground reaction forces and thruster forces as a decision variable in QP formulation. Furthermore, we will implement the controller on the hardware of Harpy, which can be challenging due to possible hardware limitations such as computing power, sensor noise, and communication delays.










\printbibliography

\end{document}